\documentclass[10pt,twocolumn,letterpaper]{article}

\usepackage{cvpr}              
\usepackage{multirow}

\usepackage[dvipsnames]{xcolor}

\definecolor{cvprblue}{rgb}{0.21,0.49,0.74}
\usepackage[pagebackref,breaklinks,colorlinks,citecolor=cvprblue]{hyperref}

\def\paperID{5125} 
\def\confName{CVPR}
\def\confYear{2024}

\title{Open-Vocabulary Segmentation with Semantic-Assisted Calibration}

\author{Yong Liu\textsuperscript{1}\footnotemark[1]~,
        Sule Bai\textsuperscript{1}\footnotemark[1]~,
        Guanbin Li\textsuperscript{2}~,
        Yitong Wang\textsuperscript{3}~,
        Yansong Tang\textsuperscript{1}\footnotemark[2]~\\
\textsuperscript{1}Shenzhen Key Laboratory of Ubiquitous Data Enabling, Shenzhen International \\Graduate School, Tsinghua University, China~
\textsuperscript{2}Sun Yat-sen University~
\textsuperscript{3}ByteDance Inc.\\
{\tt\small \{liuyong23, bsl23\}@mails.tsinghua.edu.cn, tang.yansong@sz.tsinghua.edu.cn}
}

\begin{document}
\maketitle
\footnotetext[1]{Equal contribution}
\footnotetext[2]{Corresponding author}

\begin{abstract}
This paper studies open-vocabulary segmentation (OVS) through 
calibrating in-vocabulary and domain-biased embedding space with  generalized contextual prior of CLIP. As the core of open-vocabulary understanding, alignment of visual content with the semantics of unbounded text has become the bottleneck of this field. To address this challenge, recent works propose to utilize CLIP as an additional classifier and aggregate model predictions with CLIP classification results. Despite their remarkable progress, performance of OVS methods in relevant scenarios is still unsatisfactory compared with supervised counterparts. We attribute this to the in-vocabulary embedding and domain-biased CLIP prediction. To this end, we present a Semantic-assisted CAlibration Network (SCAN). In SCAN, we incorporate generalized semantic prior of CLIP into proposal embedding to avoid collapsing on known categories. Besides, a contextual shift strategy is applied to mitigate the lack of global context and unnatural background noise.
With above designs, SCAN achieves state-of-the-art performance on all popular open-vocabulary segmentation benchmarks. 
Furthermore, we also focus on the problem of existing evaluation system that ignores semantic duplication across categories, and propose a new metric called Semantic-Guided IoU (SG-IoU). Code is available \href{https://github.com/yongliu20/SCAN}{here}.
\end{abstract}

\section{Introduction}
Semantic segmentation is one of the most fundamental tasks in computer vision, which targets at assigning semantic category to pixels in an image. Despite achieving excellent performance in recent years~\cite{sca,qdmn,unet,gsfm,segformer, deeplab, segnext, yan2023locating,qu2023learning}, traditional semantic segmentation approaches rely on predefined sets of training categories. 
Consequently, these methods falter when encountering categories absent during the training phase, significantly impeding their real-world applicability. 
Such challenge has inspired the exploration of Open-Vocabulary Segmentation (OVS) setting~\cite{zegformer,gkc,pmosr,PAD,openseg,spnet,lseg, unilseg}. 
Different from traditional closed-set segmentation, OVS methods can segment arbitrary categories given only text inputs as guidance, 
which has many potential applications such as autonomous navigation for UAV~\cite{chen1,chen2} and UGV~\cite{chen3}, mobile crowdsensing~\cite{chen4, chen5}, large scale robot swarm collaborations~\cite{chen6}.

\begin{figure}[t]
    \centering
    \includegraphics[width=\linewidth]{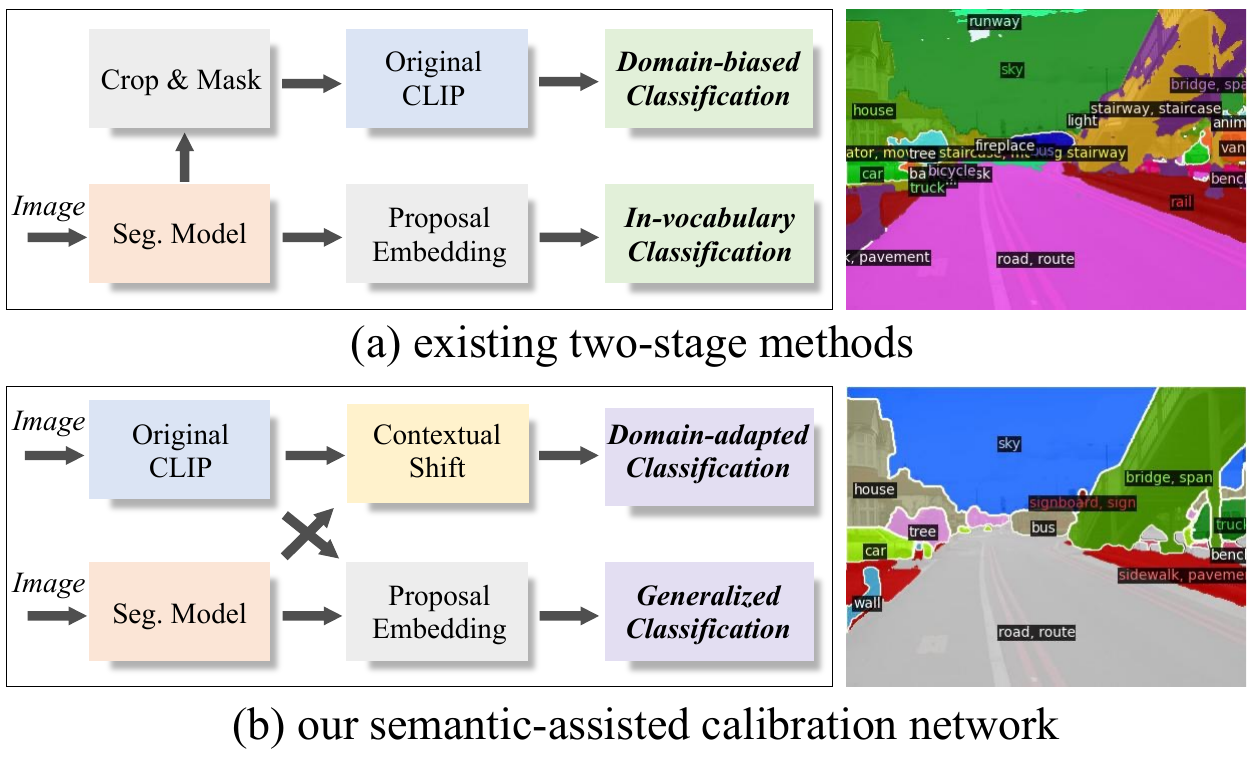}
    \vspace{-17pt}
    \caption{Illustration of existing two-stage methods and our SCAN. Limited by domain-biased CLIP classification and in-vocabulary model classification, existing methods struggle to align visual content with unbounded text.
    By incorporating generalized semantic guidance of CLIP to proposal embedding and perform contextual shift, our SCAN achieves excellent OVS performance.}
    \label{fig:teaser}
    \vspace{-12pt}
\end{figure}

It is extremely challenging to accurately identify unseen categories without the intervention of external knowledge. 
Therefore, an intuitive idea is to introduce large-scale vision-language model~\cite{clip, align} trained with numerous sources to extend the semantic space of segmentation models. 
Motivated by this, some studies~\cite{Simbaseline, zegformer, openseg, odise} adopt a two-stage pipeline. This approach first generates class-agnostic mask proposals with segmentation models, following which the pre-trained CLIP~\cite{clip} serves as an additional classifier to execute mask-level visual-linguistic alignment. The objective is to recognize open-vocabulary concepts by combining the prior knowledge of both CLIP and segmentation models.
Despite advancements  under this paradigm, its capacity to align visual content with unbounded text still falls below the anticipated outcomes considerably.
As shown in \Cref{fig:teaser} (a), we analyze this contrast mainly stems from two aspects: 1) the proposal embedding of segmentation model is turned to fit training semantic space, making segmentation model classification collapse into in-vocabulary prediction and insensitive to novel concepts. 2) the visual inputs for pre-trained CLIP have significant domain bias. Specifically, to highlight the target area and mitigate the influence from undesired regions, the input to CLIP is  sub-images after cropping and masking, which deviates significantly from the natural image distribution, \ie, the visual domain of pre-trained CLIP. 
Such bias leads to the loss of contextual guidance as well as incorrect background prior, and thus impairs the performance. 


Therefore, a natural question arises: 
how to introduce unrestricted knowledge space while mitigating domain bias caused by unnatural background and providing global context?
It occurred to us that CLIP has well-aligned visual-linguistic space and strong capability of detecting latent semantics from natural images. The \texttt{[CLS]} token embedding extracted by CLIP condenses the context of the whole image and implicitly expresses the associated semantic distribution. 
With this semantic assistance, feature space of proposal embedding and the biased visual domain of CLIP can be calibrated towards more generalized recognition.

Inspired by this, we present a Semantic-assisted CAlibration Network (SCAN). 
On the one hand, SCAN employs a semantic integration module designed to incorporate the global semantic perception of original CLIP into proposal embedding. It extends the semantic space and alleviates the potential degradation towards in-vocabulary classification. 
On the other hand, we propose a contextual shift strategy to advance the open-vocabulary recognition ability of CLIP for domain-biased images. By replacing background tokens with appropriate contextual representations, \ie, \texttt{[CLS]} embedding of whole image, this strategy mitigates domain bias at the feature level through semantic complementation.
With above designs calibrating both in-vocabulary and out-vocabulary semantic space, our SCAN achieves the best performance on all popular open-vocabulary semantic segmentation benchmarks. Extensive experiments and analysis also demonstrate the rationality of our motivation and proposed modules.

Apart from the above contribution, we also focus on the problem of current evaluation system that neglects semantic relationships among different categories.
For example, ``table'' and ``coffee table'' exist in ADE20K-150~\cite{ade20k} dataset as different class simultaneously and the model needs to accurately distinguish between them. If a model assigns ``table'' tag to a region whose ground truth label is ``coffee table", it will be considered incorrect. We believe that under open-vocabulary scenarios, correct recognition of general semantics is sufficient, and there is no need to make this level of detailed hierarchical distinction. To this end, we present a new metric called Semantic-Guided IoU (SG-IoU), which takes semantic relationships between different categories into account during IoU calculation.

Our contributions can be summarized as follows:
\begin{itemize}
    \item We present a Semantic-assisted Calibration Network (SCAN) to boost the alignment between visual content with unbounded linguistic concepts and thus improve open-vocabulary segmentation performance.
    \item We propose semantic integration module  to alleviate in-vocabulary degradation of proposal embedding assisted by original CLIP. 
    Besides, contextual shift strategy is applied to achieve domain-adapted alignment, mitigating the lack of global context and invalid background noise.
    \item We propose a new evaluation metric called Semantic-Guided  IoU (SG-IoU). It takes the semantic relationships of different categories into account, which is more compatible with the open-vocabulary setting.
    \item Our SCAN achieves state-of-the-art on all popular benchmarks with both vanilla mIoU and our proposed SG-IoU as metric. Extensive experiments are conducted to prove the effectiveness and rationality of the proposed modules.
    
\end{itemize}

\begin{figure*}[t]
    \centering
    \includegraphics[width=\textwidth]{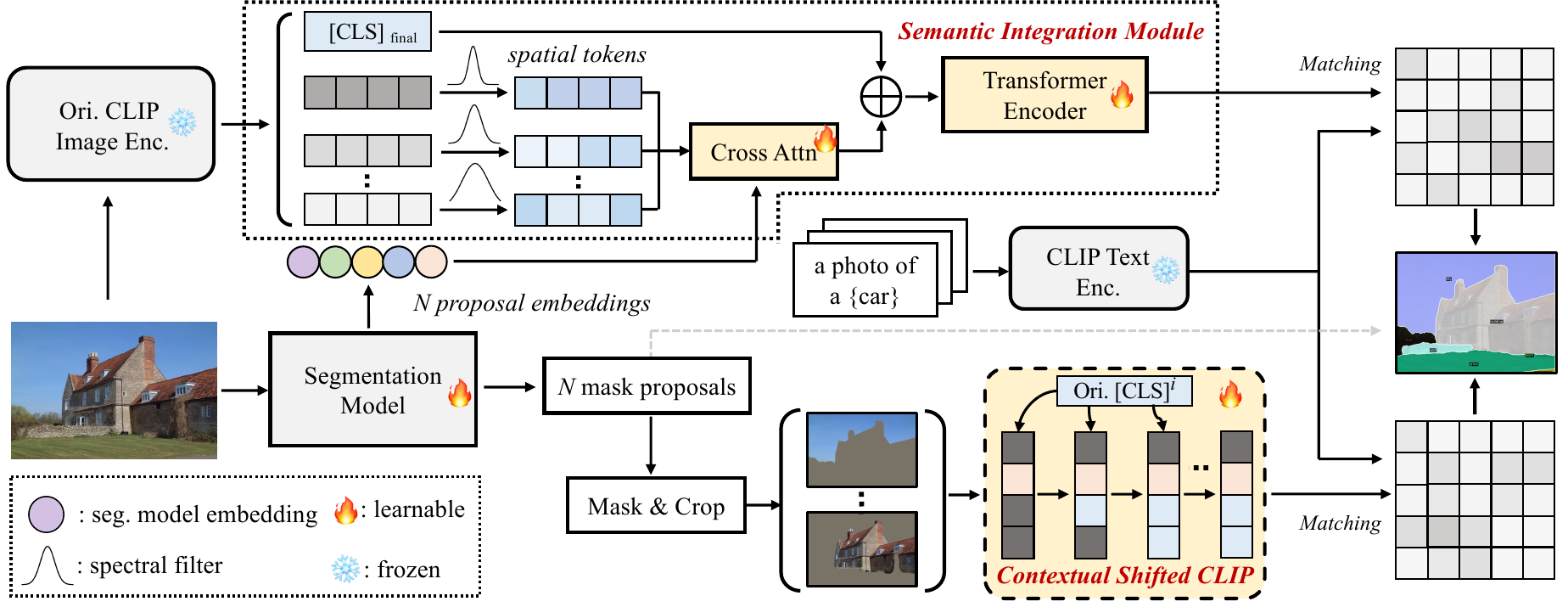}
    \caption{Pipeline of SCAN. Firstly, a segmentation model is used to generate class-agnostic masks and corresponding proposal embeddings for cross-modal alignment. To avoid collapse into known categories, the proposal embeddings are calibrated by integrating global semantic prior of CLIP in Semantic Integration Module. Besides, the cropped and masked images are input to Contextual Shifted CLIP for domain-adapted classification. Finally, the matching scores of both model embeddings and CLIP are combined to assign category labels.}
    \label{fig:pipeline}
\end{figure*}



\section{Related Work}
\paragraph{Open-Vocabulary Segmentation.}
The open-vocabulary segmentation task aims to segment an image and identify regions with arbitrary text queries~\cite{openseg,zs3net, soc, spnet}.
Pioneering work~\cite{spnet} replaces the output convolution layer by computing the similarity between visual features and linguistic embeddings, which has become common practice.
More recently, a two-stage pipeline~\cite{openseg,Simbaseline,zegformer,adapt-mask, odise} is proposed: the model first generates class-agnostic mask proposals, then 
 a pretrained CLIP~\cite{clip} is utilized to perform sub-image classification by cropping and masking corresponding regions. Afterward, the prediction of CLIP is ensembled with the classification results of segmentation model. With the combination of both in-vocabulary and out-vocabulary classification, these methods obtains excellent improvement.
Subsequently, 
SAN~\cite{san} designs a side-adapter network to leverage CLIP features for decoupling segmentation and classification.
OVSeg~\cite{adapt-mask} observes that masked background regions affect the recognization ability of CLIP due to the distribution difference. Thus, it proposes to finetune the pretrained CLIP with such images and collect a domain-biased training dataset.
Aiming to improve model efficiency, GKC~\cite{gkc} presents text-guided knowledge distillation strategy to transfer CLIP knowledge to specific classification layer.
Although above methods have made remarkable progress, they are still susceptible to bounded training semantic space due to the crucial learnable part exists in framework. To address the overfitting and domain-biased problems, we propose SCAN that calibrates both in-vocabulary and out-vocabulary space with the assistance of global semantic prior of CLIP.

\vspace{-10pt}
\paragraph{Vision-Language Pre-training.}
Vision-language pre-training aims to learn a joint visual-linguistic representation space. Limited to small-scale datasets, early approaches~\cite{pretrain1, pretrain2, pretrain3, pretrain4,qu2022siri} struggled to achieve good performance and required fine-tuning on downstream tasks. With the availability of large-scale web data, recent works~\cite{pretrain5, clip} have showed the advantages of leveraging such data to train a more robust multi-modal representation space. Among them, CLIP~\cite{clip}, the most popular vision-language method, leverages the idea of contrastive learning to connect images with their corresponding captions and has achieved impressive cross-modal alignment performance. Inspired by previous works~\cite{zegformer, lseg, Simbaseline, oclip}, we also take advantage of the well-aligned and generalised space of CLIP to enhance open-vocabulary segmenation.

\section{Method}
\Cref{fig:pipeline} shows the pipeline of SCAN. The framework follows two-stage paradigm, \ie, we first take a segmentation model~\cite{cheng2021mask2former} to generate a group of class-agnostic mask proposals $M_N\in\mathbb{R}^{N\times H\times W}$ and corresponding proposal embeddings $F_N\in\mathbb{R}^{N\times C}$, where $N$ and $C$ indicate the number of learnable queries and embedding dimension. $H$, $W$ denote the spatial size of input image.  The proposal embeddings are leveraged to align with linguistic features for model-classification. In our SCAN, a Semantic Integration Module (SIM) is proposed to transfer global semantic prior originated from CLIP into proposal embeddings $F_N$, which calibrates the model feature space to accommodate both in-vocabulary and out-vocabulary semantics. On the other hand, mask proposals $M_N$ are used to generate sub-images by cropping and masking related regions from input natural image. The processed sub-images are sent to CLIP~\cite{clip} for classification at the mask level. We propose a Contextual Shift strategy (CS) to alleviate the domain bias and noise caused by masked background pixels and improve the classification performance of CLIP for such sub-images. Finally, the classification results from both CLIP and proposal embeddings are combined for output.

\subsection{Semantic Integration Module}
The learnable proposal embedding used for model classification suffers from overfitting to training semantics and insensitive to novel categories. To relieve this problem, we propose the Semantic Integration Module (SIM). The core idea of SIM is to calibrate the semantic response of mask proposal embeddings by incorporating the prior knowledge  of CLIP~\cite{clip}. 
In SIM, we use a frozen CLIP to extract implicit semantics of the input image $I \in \mathbb{R}^{H\times W\times 3}$ and obtain the progressive features $\{F^i_{HW}, F^i_{CLS}\}$, where $F^i_{HW} \in \mathbb{R}^{\frac{H}{14}\times \frac{W}{14}\times C}$ and  $F^i_{CLS} \in \mathbb{R}^{1\times C}$ denote the output of $i\text{-}th$ layer in CLIP image encoder. To fully utilize the coarse-grained and fine-grained perception of CLIP, we introduce both the spatial tokens $F_{HW}$ and the general \texttt{[CLS]} token $F_{CLS}$ into proposal embeddings.

Considering that the purpose of feature integration is to benefit high-level semantic matching, directly interacting with spatial token embedding $F_{HW}$ may bring harmful texture noise  due to local details involved in $F_{HW}$.
Some theoretical researches~\cite{frequency1, frequency2, frequency3} about neural network from spectral domain propose that low-frequency components correspond to high-level semantic information while ignoring details.
Inspired by that, we design a simple low-frequency enhancement structure to suppress potential noise.
Take $F^i_{HW}$ as an example, the process of performing low-frequency enhancement can be represented by:
\begin{equation}
\begin{split}
    g^i &= Gaussian(h, w, \sigma),\\
    F_s^i &= FFT(F^i_{HW}) * g,\\
    F_s^i &= IFFT(ReLU(Conv(F_s^i))) + F^i_{HW},\\
\end{split}
\end{equation}
where $FFT$ and $IFFT$ are Fourier transform and Fourier inverse transform. $g^i$ denotes the filtering coefficient map with the same spatial size of the feature $F^i_{HW}$. The center of the coefficient map has the value of 0 and increases around in the form of Gaussian (without spectrum centralization, the center of the spectrum after FFT is high frequency, and the surrounding is low frequency). $\sigma$ is cutoff frequency and $*$ means element-wise product.

After performing low-frequency enhancement on selected CLIP layers, we concatenate the enhanced features to $F_s \in \mathbb{R}^{m\times\frac{H}{14}\times \frac{W}{14}\times C}$, where m denotes the number of selected CLIP layers.
Then, the content prior of $F_s$ is injected to proposal embeddings $F_N$ by multi-head cross-attention with $F_N$ as the Query and $F_s$ as the Key and Value:
\begin{equation}
    F_N' = MHA(F_N, F_s, F_s),
\end{equation}
where MHA indicates vanilla multi-head attention and $F_N' \in \mathbb{R}^{N\times C}$, $N$ is the number of learnable query.

The role of proposal embedding $F_N'$ is to align with unbounded linguistic features, while the spatial tokens in the middle layer of CLIP have not actually been transformed into the vision-language unified space. Thus, we further leverage the aligned CLIP visual embedding $F_{CLS}^{final}\in \mathbb{R}^{1\times C}$ to bridge the gap between visual and linguistic space. We add the $F_{CLS}^{final}$ to $F_N'$ with a learnable factor $\gamma$ initialized as 0.1 under the help of broadcast mechanism. Then the features are fully interacted in transformer encoder layer and generate the final aligned proposal embeddings $F_N^{final}$. This process can be formulated as:
\begin{equation}
\begin{split}
    &F_N^{final} = Trans. Enc.(F_N' + \gamma * F_{CLS}^{final}).
\end{split}
\end{equation}

\subsection{Contextual Shift}

\begin{figure}[t]
    \centering
    \includegraphics[width=\linewidth]{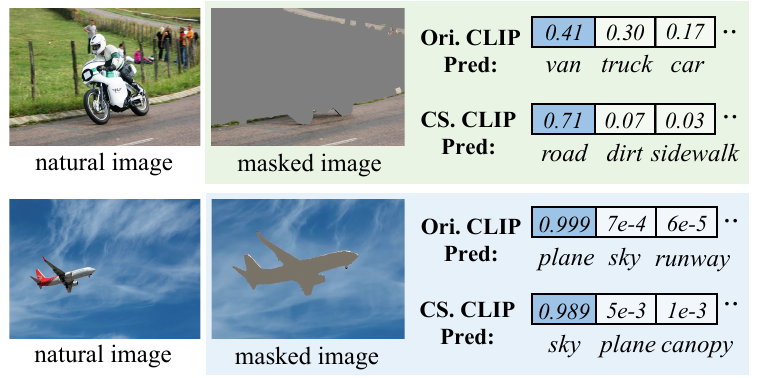}
    \vspace{-15pt}
    \caption{Illustration of image domain bias and corresponding detriment to vision-language alignment. The right side shows the classification confidence for masked images. ``Ori.CLIP" and ``CS.CLIP" demonstrate the original CLIP and our contextual shifted CLIP, respectively.}
    \label{fig:cs1}
\end{figure}

By taking pre-trained CLIP~\cite{clip} as an extra classifier, previous two-stage approaches expect to exploit the powerful generalization capability of CLIP to tackle novel classes.
But the reality is not as perfect as it seems.
As \Cref{fig:cs1} shows, the image domain has been greatly shifted from natural distribution due to the masked patches.
Such domain bias, coupled with the lack of global context, can dramatically deteriorate the recognition ability of CLIP.
Besides, the shape of masked pixels also interferes with CLIP predictions. Masking background forces corresponding regions to the same value, which imposes strong erroneous prior. For example, a plane in the sky is masked. The shape of the plane still causes related response while it is difficult to recognize the concept of ``sky" in the foreground due to its irregularity and unnatural background.

 \begin{figure}[t]
    \centering
    \includegraphics[width=\linewidth]{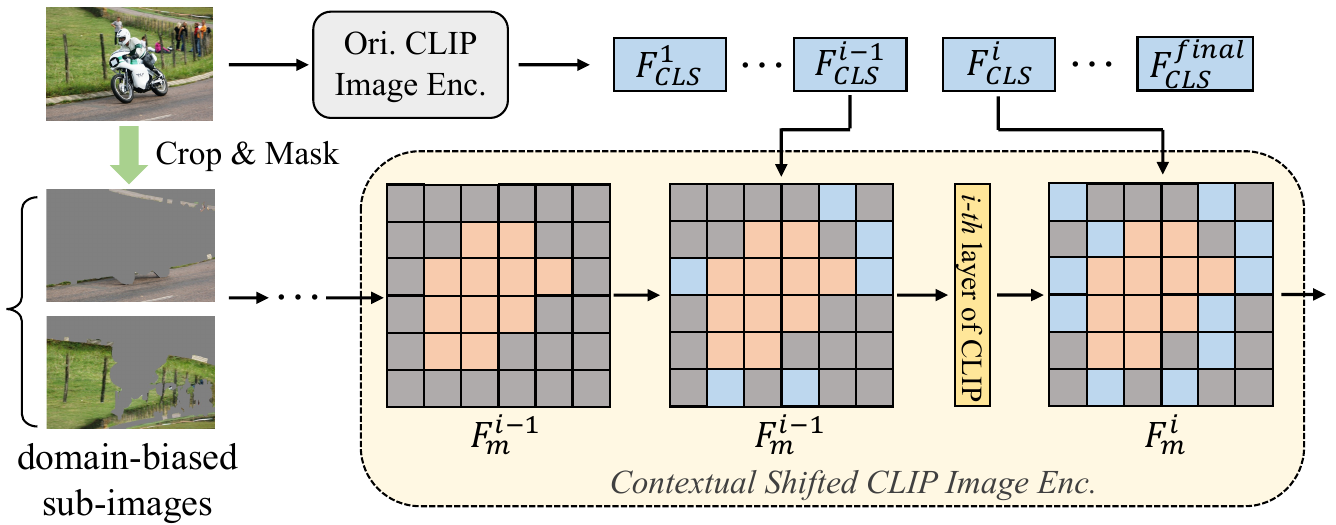}
    \vspace{-15pt}
    \caption{Process of applying contextual shift strategy.}
    \label{fig:cs2}
    \vspace{-5pt}
\end{figure}

To address this issue, we propose the Contextual Shift (CS) strategy. The key idea of CS is to replace the background token embeddings with global \texttt{[CLS]} token generated by original CLIP from whole image during the forward process. Considering the different sizes and shapes of various segmentation masks, we randomly replace a certain percentage $\alpha$ of the background areas in selected layers of CLIP. Take the $k\text{-}th$ segmentation masks as example, the vanilla forward process of CLIP for region classification is:
\begin{equation}
    F_m^i = \left\{
    \begin{array}{ll}
        \mathcal{V}^i(\delta(I, M_k)), & \text{if } i = 0 \\
        \mathcal{V}^i(F_m^{i-1}), & \text{if } i \geq 1
    \end{array}
\right.
\end{equation}
where $\mathcal{V}^i$ denotes the $i\text{-}th$ layer of CLIP visual encoder. $F_m^i$ is the output features of $i\text{-}th$ layer. 
$\delta$ indicates the crop and mask operation for generating sub-images.
$M_k$ and $I$ are segmentation mask and original input image, respectively. With the CS strategy that introducing context prior within global \texttt{[CLS]} token embedding generated by original CLIP from natural image, the updated forward process can be formulated as:
\begin{equation}
    F_m^i = \left\{
    \begin{array}{ll}
        \mathcal{V}^i(\delta(I, M_k)), & \text{if } i = 0 \\
        \mathcal{V}^i(F_m^{i-1}|(M_k, F_{CLS}^{i-1}, \alpha)), & \text{if }  i \in idx \\
        \mathcal{V}^i(F_m^{i-1}), & \text{if } others
    \end{array}
\right.
\end{equation}
where $F_{CLS}^{i-1}$ is \texttt{[CLS]} embedding generated by (i-1)$\text{-}th$ CLIP layer from natural input image. $idx$ indicates the selected replacing layers of CLIP. $(F_m^{i-1}|(M_k, F_{CLS}^{i-1}, \alpha))$ means replace $\alpha$\% of the mask background area  with the \texttt{[CLS]} embedding $F_{CLS}$ extracted from original image. The process is illustrated in \Cref{fig:cs2}.

On the one hand, the global \texttt{[CLS]} embedding obtained from natural image can provide contextual information to relieve the domain bias and aid semantic prediction.
On the other hand, such random replacing operation disrupts the shape of the background area, reducing the effect of error  distribution of consistent background pixels.
As shown in \Cref{fig:cs1}, CS strategy can greatly improve the cross-modal alignment of domain-biased images with the aforementioned advantages.
Besides, to better adapt CLIP to the shifted domain, we also follow OVSeg~\cite{adapt-mask} to finetune the contextual shifted CLIP on the masked images dataset~\cite{adapt-mask}. The dataset is collected from COCO Caption~\cite{cococaption}.


\subsection{Semantic-Guided Evaluation Metric}\label{sec:sgiou}

 \begin{figure}[t]
    \centering
    \includegraphics[width=\linewidth]{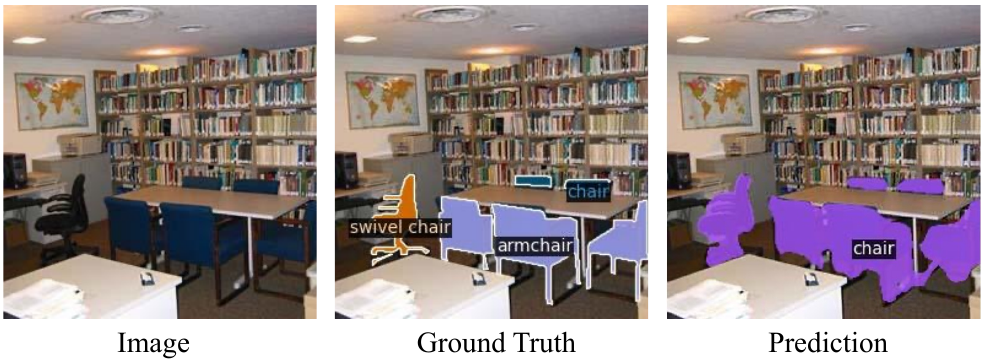}
    \caption{Explanation of potential problems exist in the current evaluation system. There exists severe semantic duplication, \ie, synonyms and parent categories, in benchmarks, while current metric does not take the semantic relationships between different categories into account.}
    \label{fig:metric}
\end{figure}

Existing OVS works tend to directly take supervised semantic segmentation benchmarks with mIoU metric for evaluation. However, we observe that such evaluation is not completely applicable to open-vocabulary settings. Specifically, there exists severe semantic duplication, \textit{i.e.}, synonyms or hypernyms, in these supervised benchmarks. 
For example, ``chair", ``armchair", and ``swivel chair" exist in the ADE20K-150 dataset~\cite{ade20k} as different class simultaneously. As \Cref{fig:metric} shows, if a model assigns  ``chair" tag to a region but the corresponding ground truth label is ``armchair", it will be considered incorrect in the existing evaluation system. 
Such category setting and evaluation is appropriate for closed-set segmentation tasks because their models are trained to distinguish between these fine-grained concepts. But for open-vocabulary segmentation setting, we argue that the responsibility of the model is to discern the correct semantic, \eg, it should also be correct if models recognize the regions belong to ``armchair" as ``chair". In addition, since the required categories are manually given under real scenarios,
users will not be inclined to give semantic duplicated categories.

Inspired by this observation, we propose to reorganize the calculation process of mIoU under existing popular benchmarks and present a new metric called Semantic-Guided IoU (SG-IoU) for open-vocabulary setting.
The core idea of SG-IoU is to take semantic relationships between different categories into account when calculating whether a prediction is consistent with the ground truth. Specifically, we manually determine the hierarchical relations among various categories and obtain a series of category semantic association matrix. When calculating the intersection between prediction and ground truth, regions predicted to be corresponding parent or synonymous classes are also taken into account. In addition, we employ a balance factor to avoid erroneous metric boosts due to the potential overfavouritism of the model to the parent categories. This factor is related to the accuracy of the parent classes.
Take $q\text{-}th$ class as an example, the calculation process can be formulated as:

 \begin{equation}
 \begin{split}
    SG\text{-}IoU(q) = \frac{P_{q}G_q + P_{Q}G_q*\beta}{P_q + G_q - P_{q}G_q}, 
    \beta = \frac{P_QG_q + P_QG_Q}{P_Q}
\end{split}
\end{equation}
where $P_QG_q$ means the predicted class is $Q$ and the ground truth category is $q$. $Q$ is the  synonyms and parent categories of $q$. $\beta$ is the balance factor. Due to limited space, please see more descriptions and demonstration of the category semantic association in supplementary materials.



\section{Experiments}

\begin{table*}[t]
\small
\centering
\renewcommand\arraystretch{1.1}
\caption{Performance comparison with state-of-the-art methods. SimSeg\dag~\cite{Simbaseline} is trained with a subset of COCO Stuff in their paper. For a fair comparison, we reproduce their method on the full COCO Stuff with their officially released code. RN101: ResNet-101~\cite{resnet}; EN-B7: EfficientNet-B7~\cite{efficientnet}. ADE, PC, and VOC denote ADE20K~\cite{ade20k}, Pascal Context~\cite{pascal}, and Pascal VOC~\cite{pascal-voc}, respectively.}
\setlength{\tabcolsep}{3.7mm}{
\begin{tabular}{l|l|c|c|c|c|c|c}
\toprule
\multirow{1}{*}{Method} & \multirow{1}{*}{VL-Model} & \multirow{1}{*}{Training Dataset} 
& ADE-150 & ADE-847 & PC-59  & PC-459  & VOC  \\
\hline
Group-VIT~\cite{groupvit}  & rand. init. & CC12M+YFCC & -    & -    & 22.4 & -   & 52.3 \\ 
\hline
LSeg+~\cite{lseg}   & ALIGN RN101 & COCO  & 13.0 & 2.5 & 36.0  & 5.2       & 59.0  \\
OpenSeg~\cite{openseg}    & ALIGN RN101  & COCO & 15.3 & 4.0  & 36.9 & 6.5       & 60.0  \\
LSeg+~\cite{lseg}  & ALIGN EN-B7 & COCO  & 18.0 & 3.8  & 46.5 & 7.8    & - \\
OpenSeg~\cite{openseg}  & ALIGN EN-B7  & COCO  & 21.1 & 6.3 & 42.1 & 9.0     & -  \\
OpenSeg~\cite{openseg}    & ALIGN EN-B7  & COCO+Loc. Narr. & 28.6 & 8.8  & 48.2 & 12.2    & 72.2 \\ 
\hline
SimSeg~\cite{Simbaseline}  & CLIP ViT-B/16    & COCO  & 20.5 & 7.0 & 47.7 & 8.7     & 88.4  \\
SimSeg\dag~\cite{Simbaseline}   & CLIP ViT-B/16 & COCO & 21.1 & 6.9  & 51.9 & 9.7     & 91.8  \\
OVSeg~\cite{adapt-mask}  & CLIP ViT-B/16  & COCO  & 24.8 & 7.1    & 53.3 & 11.0 & 92.6 \\
MAFT~\cite{maft} & CLIP ViT-B/16  & COCO  & 29.1 & 10.1 & 53.5  & 12.8  & 90.0  \\
SAN~\cite{san}    & CLIP ViT-B/16  & COCO & 27.5 &10.1   &53.8 &12.6     &94.0  \\
SCAN (Ours)  & CLIP ViT-B/16    & COCO  &\textbf{30.8} &\textbf{10.8}   &\textbf{58.4}  &\textbf{13.2}     &\textbf{97.0} \\
\hline
MaskCLIP~\cite{maskclip} & CLIP ViT-L/14    & COCO  & 23.7 & 8.2  & 45.9  & 10.0     & - \\
SimSeg\dag ~\cite{Simbaseline}     & CLIP ViT-L/14     & COCO  & 21.7 & 7.1   & 52.2   & 10.2  & 92.3  \\
OVSeg~\cite{adapt-mask}    & CLIP ViT-L/14   & COCO & 29.6 & 9.0 & 55.7 & 12.4   & 94.5 \\
ODISE~\cite{odise}  & CLIP ViT-L/14   & COCO & 29.9 & 11.1 & 57.3 & 14.5   & -\\
SAN~\cite{san}  & CLIP ViT-L/14    & COCO  &32.1 & 12.4   &57.7  &15.7     &94.6 \\
SCAN (Ours)  & CLIP ViT-L/14    & COCO  &\textbf{33.5} &\textbf{14.0}   &\textbf{59.3}  &\textbf{16.7}     &\textbf{97.2} \\
                \bottomrule
\end{tabular}
}
\label{tab:main_result}
\vspace{-10pt}
\end{table*}

\subsection{Datasets and Evaluation Metrics}
\textbf{Training Dataset} Following previous works~\cite{adapt-mask, Simbaseline, san, zegformer, maft}, we train the segmentation model of our SCAN on COCO-Stuff~\cite{coco} with 171 categories. CLIP~\cite{clip} for mask-level classification is finetuned on masked images dataset proposed by OVSeg~\cite{adapt-mask}. The dataset is collected from COCO Captions~\cite{cococaption}.

\noindent \textbf{Evaluation Dataset}
To evaluate the effectiveness of our method, we conduct extensive experiments on the popular image benchmarks, ADE20K150~\cite{ade20k}, ADE20K847~\cite{ade20k}, Pascal VOC~\cite{pascal-voc}, Pascal Context-59~\cite{pascal}, and Pascal Context-459~\cite{pascal}.

ADE20K is a large-scale scene understanding benchmark, containing 20k training images, 2k validation images, and 3k testing images. 
There are two splits of this dataset. 
ADE20K-150 contains 150 semantic classes whereas ADE20K-847 has 847 classes. The images of both are the same.
Pascal Context is an extension of Pascal VOC 2010, containing 4,998 training images and 5,005 validation images. 
We take the commonly used PC-59 and challenging PC-459 version for validation.
Pascal VOC contains 11,185 training images and 1,449 validation images from 20 classes. We use the provided augmented annotations.

\noindent \textbf{Evaluation Metric} Following previous works~\cite{Simbaseline,zegformer,openseg}, we take the \textit{mean-intersection-over-union} (mIoU) as the metric to compare our model with previous state-of-the-art methods. In addition, we also report the corresponding results measured by our proposed SG-IoU. 

\subsection{Implementation Details}
For segmentation model, our implementation is based on {\tt detectron2}~\cite{wu2019detectron2}.
All image-based models are trained with batch size of 32 and training iteration of 120k. The base learning rate is 0.00006 with a polynomial schedule. The shortest edge of input image is resized to 640. For data augmentation, random flip and color augmentation are adopted. The weight decay of the segmentation model is 0.01. 
The backbone of segmentation model is Swin Transformer-Base~\cite{swin}.
The CLIP~\cite{clip} version is ViT-L/14, implemented by OpenCLIP.
For the weights of the loss function, we set 5 and  2 for segmentation loss and classification loss, respectively. The segmentation loss consists of dice loss and cross entropy loss. The classification loss is cross entropy loss.  Other hyperparameters are the same as Mask2Former~\cite{cheng2021mask2former}.
For fine-tuned CLIP, the training process is the same as OVSeg~\cite{adapt-mask}. 

\subsection{Main Results}
We compare our model with existing state-of-the-art approaches in \Cref{tab:main_result}. To make it clear, we group the methods according to the utilized vision-language model and report the performance of our SCAN with CLIP ViT-B/16 as well as ViT-L/14~\cite{clip}. It can be seen that with global distribution prior, our model achieves the best performance on all popular benchmarks under both ViT-B and ViT-L. With ViT-B/16, our model reaches 30.8 and 58.4 on ADE-150 and PC-59, surpassing previous methods by a large margin. For ViT-L/14, our SCAN overpasses previous state-of-the-art by about 1.5\% on ADE-150 and ADE-847. On PC-59 and PC-459, SCAN achieves 59.3 and 16.7, respectively. 

\subsection{Evaluation with SG-IoU}
The above comparisons are based on vanilla evaluation system. As explained in \Cref{sec:sgiou}, there exists problems when directly using traditional mIoU as evaluation metric for open-vocabulary segmentation performance under existing datasets. Therefore, we also report the results evaluated with the proposed SG-IoU in \Cref{tab:sgiou}. By taking semantic relationships between different categories into account, the performance  would improve and the gap between various methods is also different from \Cref{tab:main_result}. More analysis please see the supplementary materials.


\begin{table}[t]
\small
\centering

\caption{Evaluation with SG-IoU as metric of our SCAN and some open-source methods. For the sake of comparison, we report the results of vanilla mIoU in \textcolor{gray}{gray} color.}
\vspace{-7pt}
\setlength{\tabcolsep}{3.1mm}{
\begin{tabular}{l|c|c|c}
\hline
\multirow{1}{*}{Method}
& ADE-150 & ADE-847 & PC-459  \\
\hline
SimSeg\cite{Simbaseline}        & 22.6 / \textcolor{gray}{20.5} & 8.1 / \textcolor{gray}{7.0}   & 9.3 / \textcolor{gray}{8.7}    \\
OVSeg~\cite{adapt-mask}     & 30.5 / \textcolor{gray}{29.6} & 9.5 / \textcolor{gray}{9.0} & 12.7 / \textcolor{gray}{12.4} \\
MAFT~\cite{maft}    & 30.3 / \textcolor{gray}{29.1} & 11.5 / \textcolor{gray}{10.1} & 13.4 / \textcolor{gray}{12.8} \\
SAN~\cite{san}    &33.7 / \textcolor{gray}{32.1} & 13.2 / \textcolor{gray}{12.4}  &16.2 / \textcolor{gray}{15.7}   \\
\hline
SCAN(Ours)  &\textbf{34.2} / \textcolor{gray}{33.5} &\textbf{14.6} / \textcolor{gray}{14.0}   &\textbf{17.2} / \textcolor{gray}{16.7}   \\
\hline
\end{tabular}
}
\label{tab:sgiou}
\vspace{-5pt}
\end{table}

\subsection{Ablation Study}
\paragraph{Model Component Analysis} \Cref{tab:component} shows the ablation study about the proposed Semantic Integration Module (SIM) and Contextual Shift (CS) strategy. It can be seen that when using SIM or CS alone, the model performance improves because of the injection of CLIP global semantic prior. 
When using both of them, the model achieves the best performance on all benchmarks.

\begin{table}
    \centering
    \small
    \caption{Ablation experiments on the proposed modules.}
    \vspace{-7pt}
    \setlength{\tabcolsep}{13.5pt}
    \begin{tabular}{l|ccc}
    \hline
         Method  &ADE-150  &ADE-847  &PC-59 \\
         \hline
         Baseline  &31.5  &11.4  &57.1 \\
         + SIM  &32.9  &13.3  &58.6 \\
         + CS  & 32.8  &12.6  &58.3 \\
         + Both  &\textbf{33.5}  &\textbf{14.0 } &\textbf{59.3} \\
         \hline
    \end{tabular}
    
    \label{tab:component}
\end{table}

\vspace{-10pt}
\paragraph{Analysis on Semantic Integration Module} 

\Cref{tab:sim} reports the influence of different structure design of SIM. 
To verify the effectiveness precisely, corresponding experiments are conducted without contextual shift strategy.
It can be seen that without any semantic guidance, the model performs poorly. Since the final \texttt{[CLS]} embedding contains rich semantic information and has been aligned with textual domain, incorporating it to proposal embeddings can greatly prevent overfitting to training categories and improve final performance. Besides, the spatial tokens from middle layer of CLIP also contain local semantic perceptions and contribute to open-vocabulary segmentation. But due to the potential high-frequency noise, the employed spectral enhancement operation helps to  fully utilize such fine-grained semantic source. 
SIM achieves the best performance when using all of them.

\begin{table}
    \centering
    \small
    
    \caption{Different structure design of Semantic Integration Module(SIM). $F_{CLS}^{final}$, $F_{HW}$ and LFE denote leveraging the final \texttt{[CLS]} embedding, selected spatial token embeddings, and low-frequency enhancement strategy, respectively.}
    \vspace{-5pt}
    \setlength{\tabcolsep}{5.8pt}
    \begin{tabular}{ccc|ccc}
    \hline
         $F_{CLS}^{final}$ & $F_{HW}$  & LFE  &ADE-150  &ADE-847  &PC-59 \\
         \hline
         &  &  &31.5  &11.4  &57.1 \\
         \checkmark &  &  &32.5 &12.7  &57.5 \\
         &\checkmark  &  &32.0  &11.9  &57.5 \\
         \checkmark&\checkmark  &  &32.6  &13.0  &58.1 \\
         &\checkmark  &\checkmark  &32.4  &12.2  &57.7 \\
         \checkmark &\checkmark  &\checkmark  &\textbf{32.9}  &\textbf{13.3}  &\textbf{58.6} \\
         \hline
    \end{tabular}
    \label{tab:sim}
\end{table}

    

We also conduct experiments about the selected layer of spatial token embeddings from original CLIP and the selection of frequency kernels, \ie, the cutoff frequency $\sigma$. 
The results are shown in \Cref{tab:lfm}. We can see that the spatial embeddings should hold information of multi-level granularity rather than all from deep layers. We attribute this to the complementary nature of the multi-granularity information. Besides, an appropriate cut-off frequency is also required due to the presence of undesirable texture noise.
\begin{table}
    \centering
    \small
    \setlength{\tabcolsep}{8pt}
    \caption{Selected CLIP layers and frequency kernels for SIM.}
    \vspace{-7pt}
    \begin{tabular}{lc|ccc}
    \hline
         Layers & Kernel  &ADE-150 &ADE-847 &PC-59  \\
         \hline
         15, 18, 21 &7, 5, 3  &32.5 &12.9 &58.3  \\
         15, 18, 21 &9, 7, 3  &32.4 &13.0 &58.5  \\
         12, 18, 24 &7, 5, 3  &33.3 &13.6 &58.9  \\
         12, 18, 24 &9, 7, 3  &\textbf{33.5} &\textbf{14.0} &\textbf{59.3}  \\
         12, 18, 24 &11, 9, 7  &33.0 &13.4 &58.4  \\
         \hline
    \end{tabular}
    
    \label{tab:lfm}
\end{table}

\begin{figure*}[t]
    \centering
    \includegraphics[width=\textwidth]{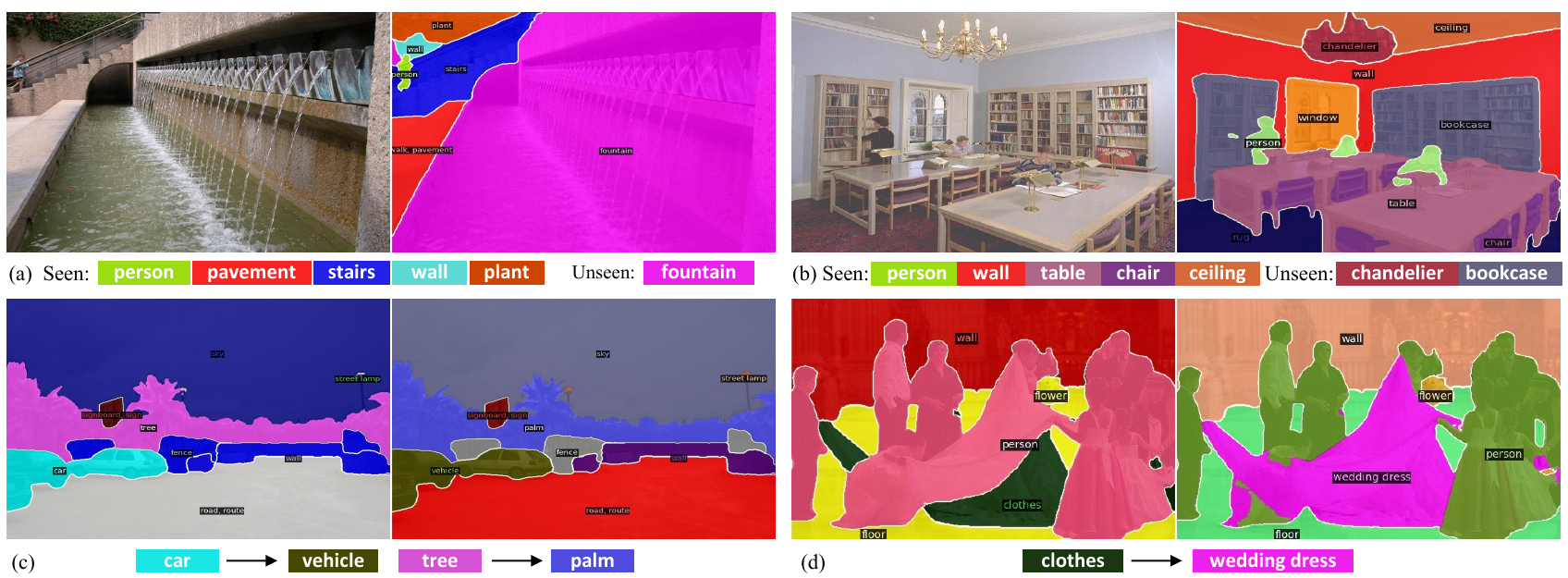}
    \caption{Visualization segmentation results. (a) and (b) demonstrate the excellent segmentation of our method for seen and unseen categories. (c) and (d) displays the adaptability for flexible text query. Best viewed in color.}
    \label{fig:visual}
    \vspace{-10pt}
\end{figure*}

\vspace{-10pt}
\paragraph{Contextual Shift Strategy} 
Here we compare different source of substitution tokens for contextual shift strategy as well as the influence of related location and ratios.

\textit{Sources of substitutions:} to relieve the problems of domain bias, we randomly replace patch embeddings belong to background area with \texttt{[CLS]} embedding from corresponding layers of original CLIP with natural image as input. We also experiment with other replacing strategies under the same replacement percentage in \Cref{tab:cs}. 
Specifically, we trial with utilizing random noise to replace (Random noise), randomly preserving the original pixels (Original background),  and with learnable tokens (Learnable prompt)~\cite{adapt-mask}. Results show that although such strategies can also disrupt the background shape, they struggles to simultaneously maintain contextual semantics and mitigate domain bias, leading to performance degradation. Besides, the learnable prompt is concerned about overfitting to seen semantics. The mIoU of taking learnable prompt is increased on ADE-150 and PC-59 datasets, which are similar to training space. But the performance drops on the more challenging benchmark ADE-847.

\begin{table}
    \centering
    \small
    \setlength{\tabcolsep}{8.1pt}
    \caption{Comparison of different substitution sources.}
    \vspace{-7pt}
    \begin{tabular}{l|ccc}
    \hline
         Method &ADE-150  &ADE-847  &PC-59 \\
         \hline
         None &32.7 &13.3 &58.6 \\
         Random noise &32.3  &13.0  &57.8  \\
         Original background &32.3  &13.1  &58.0  \\
         Learnable prompt &33.1  &13.2  &58.8  \\
         SCAN(Ours) &\textbf{33.5}  &\textbf{14.0}  &\textbf{59.3}  \\
         \hline
    \end{tabular}
    
    \label{tab:cs}
    \vspace{-10pt}
\end{table}

\textit{Location and ratios of substitutions:} \Cref{tab:cslayer} presents the results of replacing background patches at different layers of CLIP with different ratios.
It can be seen that the replacing should occur at shallow layers. If the contextual shift occurs after $11\text{-}th$ layer, the performance would drops. We analyze this is because the biased background region is already sensed by the shallow layers of CLIP and brings erroneous prior.
For replacing ratios, we find that excessively high proportion of replacement would make global context impair local area judgements, as shown in Layers of (1, 3, 5, 7, 9) with 40\% replacement ratios.
The results of layers (1, 2, 3, 4, 5) with 30\% replacement ratios also drops on ADE-150~\cite{ade20k}. We posit that this can be attributed to the fact that successive replacement tends to induce overly global representation effects, which are akin to high replacement ratios.
    

\begin{table}[t]
    \centering
    \small
    \setlength{\tabcolsep}{3.4pt}
    \caption{Replacement ratios and layers within  CS strategy.}
    \begin{tabular}{lc|ccc}
    \hline
         Layers &Ratios  &ADE-150 & ADE-847 & PC-59 \\
         \hline
         W/o Contexutal Shift  &-  &32.9 & 13.3 & 58.6   \\
         \hline
         1, 3, 5, 7, 9 &10\%  &\textbf{33.5} & 13.6 & 59.4   \\
         1, 3, 5, 7, 9 &20\%  &33.1 & 13.8 & \textbf{59.5}   \\
         1, 3, 5, 7, 9 &30\%  &\textbf{33.5} & \textbf{14.0} & 59.3   \\
         1, 3, 5, 7, 9 &40\%  &32.9 & 13.6 & 59.3   \\
         1, 3, 5, 7 &30\%  &32.8 & 13.6 & 59.0   \\
         1, 2, 3, 4, 5 &30\%  &32.2 & 13.2 & 58.9   \\
         11, 13, 15, 17, 19 &30\%  &32.2 & 12.0 & 59.1   \\
         \hline
    \end{tabular}
    
    \label{tab:cslayer}
    \vspace{-10pt}
\end{table}

\vspace{-10pt}
\paragraph{Improvement of In-Vocabulary and Domain-Biased Embedding} To prove the proposed SIM and CS strategy can relieve the overfitting problem of proposal embeddings and image domain bias for CLIP, we test the performance gains of using them alone. From \Cref{tab:only} we can see that by calibrating corresponding space, the cross-modal alignment of both mask embedding and CLIP embedding have been remarkably improved. 
With both calibration on mask embedding and CLIP embedding, performance is more significantly improved.
\begin{table}[t]
    \small
    \centering
    \caption{Improvement of model proposal embedding and domain-biased CLIP classification, respectively.}
    \vspace{-5pt}
    \renewcommand\tabcolsep{3.3pt}
    \begin{tabular}{l|c|c|c}
    \hline
        &ADE-150 &ADE-847 &PC-59\\ \hline     
    \multicolumn{4}{l}{ \textit{(a) Baseline}} \\ \hline
    Only Mask Embedding    &24.8 &9.3     &56.7  \\   
    Only CLIP Embedding  &27.7 &10.0     &48.2  \\   
    Both   &31.5 &11.4     &57.1   \\
    \hline
    \multicolumn{4}{l}{ \textit{(b) Our SCAN}} \\ \hline
    Only Mask Embedding    &26.3$\uparrow(\textbf{1.5})$ &11.1$\uparrow(\textbf{1.8})$     &57.9$\uparrow(\textbf{1.2})$   \\   
    Only CLIP Embedding    &28.9$\uparrow(\textbf{1.2})$ &10.4$\uparrow(\textbf{0.4})$     &49.9$\uparrow(\textbf{1.7})$   \\   
    Both   &33.5$\uparrow(\textbf{2.0})$ &14.0$\uparrow(\textbf{2.6})$     &59.3$\uparrow(\textbf{2.2})$  \\   
    \hline
    \end{tabular}
    \label{tab:only}
    \vspace{-10pt}
\end{table}

\subsection{Visualization}
\Cref{fig:visual} shows some segmentation cases of our SCAN. It can be seen that our method achieves excellent segmentation performance on various scenarios. Specifically, (a) and (b) demonstrate the excellent segmentation of our method for seen and unseen categories. (c) and (d) display the adaptability for flexible text query, \eg, change ``tree" to ``palm" and ``clothes" to ``wedding dress".

\section{Conclusion}
We present a Semantic-assisted CAlibration Network (SCAN) in this paper to boost vision-language alignment performance. 
In SCAN, SIM is proposed to calibrate the mask proposal embedding and relieve the overfitting problem. To compensate global context and mitigate the image domain bias, CS strategy is adopted for CLIP prediction. Extensive experiments show that SCAN achieves state-of-the-art performance on all popular open-vocabulary segmentation benchmarks. 
Besides, we focus on the problem of existing evaluation system that neglects relationships across classes, and propose a new metric called SG-IoU.

\noindent \textbf{Acknowledgements.} This work was supported in part by the National Natural Science Foundation of China under Grant 62206153 and NO.~62322608, Shenzhen Key Laboratory of Ubiquitous Data Enabling (Grant No. ZDSYS20220527171406015).

{
    \small
    \bibliographystyle{ieeenat_fullname}
    \bibliography{main}

\begin{thebibliography}{54}
\providecommand{\natexlab}[1]{#1}
\providecommand{\url}[1]{\texttt{#1}}
\expandafter\ifx\csname urlstyle\endcsname\relax
  \providecommand{\doi}[1]{doi: #1}\else
  \providecommand{\doi}{doi: \begingroup \urlstyle{rm}\Url}\fi

\bibitem[Bucher et~al.(2019)Bucher, Vu, Cord, and P{\'{e}}rez]{zs3net}
Maxime Bucher, Tuan{-}Hung Vu, Matthieu Cord, and Patrick P{\'{e}}rez.
\newblock Zero-shot semantic segmentation.
\newblock In \emph{NeurIPS}, 2019.

\bibitem[Chen et~al.(2018)Chen, Papandreou, Kokkinos, Murphy, and Yuille]{deeplab}
Liang{-}Chieh Chen, George Papandreou, Iasonas Kokkinos, Kevin Murphy, and Alan~L. Yuille.
\newblock Deeplab: Semantic image segmentation with deep convolutional nets, atrous convolution, and fully connected crfs.
\newblock \emph{TPAMI}, 2018.

\bibitem[Chen et~al.(2015{\natexlab{a}})Chen, Fang, Lin, Vedantam, Gupta, Doll{\'a}r, and Zitnick]{cococaption}
Xinlei Chen, Hao Fang, Tsung-Yi Lin, Ramakrishna Vedantam, Saurabh Gupta, Piotr Doll{\'a}r, and C~Lawrence Zitnick.
\newblock Microsoft coco captions: Data collection and evaluation server.
\newblock \emph{arXiv preprint arXiv:1504.00325}, 2015{\natexlab{a}}.

\bibitem[Chen et~al.(2015{\natexlab{b}})Chen, Purohit, Dominguez, Carpin, and Zhang]{chen2}
Xinlei Chen, Aveek Purohit, Carlos~Ruiz Dominguez, Stefano Carpin, and Pei Zhang.
\newblock Drunkwalk: Collaborative and adaptive planning for navigation of micro-aerial sensor swarms.
\newblock In \emph{Proceedings of the 13th ACM Conference on Embedded Networked Sensor Systems}, pages 295--308, 2015{\natexlab{b}}.

\bibitem[Chen et~al.(2022)Chen, Wang, Li, Ding, Dang, Wu, and Chen]{chen4}
Xuecheng Chen, Haoyang Wang, Zuxin Li, Wenbo Ding, Fan Dang, Chengye Wu, and Xinlei Chen.
\newblock Deliversense: Efficient delivery drone scheduling for crowdsensing with deep reinforcement learning.
\newblock In \emph{Adjunct Proceedings of the 2022 ACM International Joint Conference on Pervasive and Ubiquitous Computing and the 2022 ACM International Symposium on Wearable Computers}, pages 403--408, 2022.

\bibitem[Chen et~al.(2019)Chen, Li, Yu, Kholy, Ahmed, Gan, Cheng, and Liu]{pretrain1}
Yen{-}Chun Chen, Linjie Li, Licheng Yu, Ahmed~El Kholy, Faisal Ahmed, Zhe Gan, Yu Cheng, and Jingjing Liu.
\newblock {UNITER:} learning universal image-text representations.
\newblock \emph{arXiv preprint arXiv:1909.11740}, 2019.

\bibitem[Cheng et~al.(2022)Cheng, Misra, Schwing, Kirillov, and Girdhar]{cheng2021mask2former}
Bowen Cheng, Ishan Misra, Alexander~G. Schwing, Alexander Kirillov, and Rohit Girdhar.
\newblock Masked-attention mask transformer for universal image segmentation.
\newblock In \emph{CVPR}, 2022.

\bibitem[Ding et~al.(2022{\natexlab{a}})Ding, Xue, Xia, and Dai]{zegformer}
Jian Ding, Nan Xue, Gui{-}Song Xia, and Dengxin Dai.
\newblock Decoupling zero-shot semantic segmentation.
\newblock In \emph{CVPR}, 2022{\natexlab{a}}.

\bibitem[Ding et~al.(2022{\natexlab{b}})Ding, Wang, and Tu]{maskclip}
Zheng Ding, Jieke Wang, and Zhuowen Tu.
\newblock Open-vocabulary panoptic segmentation with maskclip.
\newblock \emph{arXiv preprint arXiv:2208.08984}, 2022{\natexlab{b}}.

\bibitem[Everingham et~al.(2015)Everingham, Eslami, Van~Gool, Williams, Winn, and Zisserman]{pascal-voc}
Mark Everingham, SM~Ali Eslami, Luc Van~Gool, Christopher~KI Williams, John Winn, and Andrew Zisserman.
\newblock The pascal visual object classes challenge: A retrospective.
\newblock \emph{IJCV}, 111:\penalty0 98--136, 2015.

\bibitem[Fang et~al.(2023)Fang, Zhu, Cheng, Liu, Wei, and Zhao]{yan2023locating}
Yan Fang, Feng Zhu, Bowen Cheng, Luoqi Liu, Yunchao Wei, and Yao Zhao.
\newblock Locating noise is halfway denoising for semi-supervised segmentatio.
\newblock In \emph{Proceedings of the IEEE/CVF International Conference on Computer Vision}, 2023.

\bibitem[Ghiasi et~al.(2021)Ghiasi, Gu, Cui, and Lin]{openseg}
Golnaz Ghiasi, Xiuye Gu, Yin Cui, and Tsung{-}Yi Lin.
\newblock Open-vocabulary image segmentation.
\newblock \emph{arXiv preprint arXiv:2112.12143}, 2021.

\bibitem[Guo et~al.(2022)Guo, Lu, Hou, Liu, Cheng, and Hu]{segnext}
Meng{-}Hao Guo, Chengze Lu, Qibin Hou, Zheng{-}Ning Liu, Ming{-}Ming Cheng, and Shi{-}Min Hu.
\newblock Segnext: Rethinking convolutional attention design for semantic segmentation.
\newblock \emph{arXiv preprint arXiv:2209.08575}, 2022.

\bibitem[Han et~al.(2023)Han, Liu, Liew, Ding, Liu, Wang, Tang, Yang, Feng, Zhao, et~al.]{gkc}
Kunyang Han, Yong Liu, Jun~Hao Liew, Henghui Ding, Jiajun Liu, Yitong Wang, Yansong Tang, Yujiu Yang, Jiashi Feng, Yao Zhao, et~al.
\newblock Global knowledge calibration for fast open-vocabulary segmentation.
\newblock In \emph{Proceedings of the IEEE/CVF International Conference on Computer Vision}, pages 797--807, 2023.

\bibitem[He et~al.(2016)He, Zhang, Ren, and Sun]{resnet}
Kaiming He, Xiangyu Zhang, Shaoqing Ren, and Jian Sun.
\newblock Deep residual learning for image recognition.
\newblock In \emph{CVPR}, pages 770--778, 2016.

\bibitem[He et~al.(2023)He, Ding, and Jiang]{PAD}
Shuting He, Henghui Ding, and Wei Jiang.
\newblock Primitive generation and semantic-related alignment for universal zero-shot segmentation.
\newblock In \emph{CVPR}, 2023.

\bibitem[Huang et~al.(2023)Huang, Wang, Tang, Zhang, Hu, Lu, Wang, and Liu]{sca}
Xiaoke Huang, Jianfeng Wang, Yansong Tang, Zheng Zhang, Han Hu, Jiwen Lu, Lijuan Wang, and Zicheng Liu.
\newblock Segment and caption anything.
\newblock \emph{arXiv preprint arXiv:2312.00869}, 2023.

\bibitem[Jia et~al.(2021{\natexlab{a}})Jia, Yang, Xia, Chen, Parekh, Pham, Le, Sung, Li, and Duerig]{pretrain5}
Chao Jia, Yinfei Yang, Ye Xia, Yi{-}Ting Chen, Zarana Parekh, Hieu Pham, Quoc~V. Le, Yun{-}Hsuan Sung, Zhen Li, and Tom Duerig.
\newblock Scaling up visual and vision-language representation learning with noisy text supervision.
\newblock In \emph{ICML}, 2021{\natexlab{a}}.

\bibitem[Jia et~al.(2021{\natexlab{b}})Jia, Yang, Xia, Chen, Parekh, Pham, Le, Sung, Li, and Duerig]{align}
Chao Jia, Yinfei Yang, Ye Xia, Yi-Ting Chen, Zarana Parekh, Hieu Pham, Quoc Le, Yun-Hsuan Sung, Zhen Li, and Tom Duerig.
\newblock Scaling up visual and vision-language representation learning with noisy text supervision.
\newblock In \emph{ICML}, pages 4904--4916, 2021{\natexlab{b}}.

\bibitem[Jian et~al.(2023)Jian, Liu, Shao, Wang, Chen, and Liang]{chen3}
Zhuozhu Jian, Zejia Liu, Haoyu Shao, Xueqian Wang, Xinlei Chen, and Bin Liang.
\newblock Path generation for wheeled robots autonomous navigation on vegetated terrain.
\newblock \emph{IEEE Robotics and Automation Letters}, 2023.

\bibitem[Jiao et~al.(2023)Jiao, Wei, Wang, Zhao, and Shi]{maft}
Siyu Jiao, Yunchao Wei, Yaowei Wang, Yao Zhao, and Humphrey Shi.
\newblock Learning mask-aware clip representations for zero-shot segmentation.
\newblock \emph{arXiv preprint arXiv:2310.00240}, 2023.

\bibitem[Li et~al.(2022{\natexlab{a}})Li, Weinberger, Belongie, Koltun, and Ranftl]{lseg}
Boyi Li, Kilian~Q. Weinberger, Serge~J. Belongie, Vladlen Koltun, and Ren{\'{e}} Ranftl.
\newblock Language-driven semantic segmentation.
\newblock In \emph{ICLR}, 2022{\natexlab{a}}.

\bibitem[Li et~al.(2020{\natexlab{a}})Li, Duan, Fang, Gong, and Jiang]{pretrain2}
Gen Li, Nan Duan, Yuejian Fang, Ming Gong, and Daxin Jiang.
\newblock Unicoder-vl: {A} universal encoder for vision and language by cross-modal pre-training.
\newblock In \emph{AAAI}, 2020{\natexlab{a}}.

\bibitem[Li et~al.(2022{\natexlab{b}})Li, Huang, Zhu, Tang, Li, Zhou, and Lu]{oclip}
Wanhua Li, Xiaoke Huang, Zheng Zhu, Yansong Tang, Xiu Li, Jie Zhou, and Jiwen Lu.
\newblock Ordinalclip: Learning rank prompts for language-guided ordinal regression.
\newblock \emph{NeurIPS}, pages 35313--35325, 2022{\natexlab{b}}.

\bibitem[Li et~al.(2020{\natexlab{b}})Li, Yin, Li, Zhang, Hu, Zhang, Wang, Hu, Dong, Wei, Choi, and Gao]{pretrain3}
Xiujun Li, Xi Yin, Chunyuan Li, Pengchuan Zhang, Xiaowei Hu, Lei Zhang, Lijuan Wang, Houdong Hu, Li Dong, Furu Wei, Yejin Choi, and Jianfeng Gao.
\newblock Oscar: Object-semantics aligned pre-training for vision-language tasks.
\newblock In \emph{ECCV}, 2020{\natexlab{b}}.

\bibitem[Li et~al.(2022{\natexlab{c}})Li, Man, Chen, Zhao, Wu, and Chen]{chen5}
Zuxin Li, Fanhang Man, Xuecheng Chen, Baining Zhao, Chenye Wu, and Xinlei Chen.
\newblock Tract: Towards large-scale crowdsensing with high-efficiency swarm path planning.
\newblock In \emph{Adjunct Proceedings of the 2022 ACM International Joint Conference on Pervasive and Ubiquitous Computing and the 2022 ACM International Symposium on Wearable Computers}, pages 409--414, 2022{\natexlab{c}}.

\bibitem[Liang et~al.(2022)Liang, Wu, Dai, Li, Zhao, Zhang, Zhang, Vajda, and Marculescu]{adapt-mask}
Feng Liang, Bichen Wu, Xiaoliang Dai, Kunpeng Li, Yinan Zhao, Hang Zhang, Peizhao Zhang, Peter Vajda, and Diana Marculescu.
\newblock Open-vocabulary semantic segmentation with mask-adapted {CLIP}.
\newblock \emph{arXiv preprint arXiv:2210.04150}, 2022.

\bibitem[Lin et~al.(2014)Lin, Maire, Belongie, Hays, Perona, Ramanan, Doll{\'{a}}r, and Zitnick]{coco}
Tsung{-}Yi Lin, Michael Maire, Serge~J. Belongie, James Hays, Pietro Perona, Deva Ramanan, Piotr Doll{\'{a}}r, and C.~Lawrence Zitnick.
\newblock Microsoft {COCO:} common objects in context.
\newblock In \emph{ECCV}, 2014.

\bibitem[Liu et~al.(2022{\natexlab{a}})Liu, Yu, Wang, Zhao, Wang, Tang, and Yang]{gsfm}
Yong Liu, Ran Yu, Jiahao Wang, Xinyuan Zhao, Yitong Wang, Yansong Tang, and Yujiu Yang.
\newblock Global spectral filter memory network for video object segmentation.
\newblock In \emph{ECCV}, pages 648--665, 2022{\natexlab{a}}.

\bibitem[Liu et~al.(2022{\natexlab{b}})Liu, Yu, Yin, Zhao, Zhao, Xia, and Yang]{qdmn}
Yong Liu, Ran Yu, Fei Yin, Xinyuan Zhao, Wei Zhao, Weihao Xia, and Yujiu Yang.
\newblock Learning quality-aware dynamic memory for video object segmentation.
\newblock In \emph{ECCV}, pages 468--486, 2022{\natexlab{b}}.

\bibitem[Liu et~al.(2023)Liu, Zhang, Wang, Wang, Yang, and Tang]{unilseg}
Yong Liu, Cairong Zhang, Yitong Wang, Jiahao Wang, Yujiu Yang, and Yansong Tang.
\newblock Universal segmentation at arbitrary granularity with language instruction.
\newblock \emph{arXiv preprint arXiv:2312.01623}, 2023.

\bibitem[Liu et~al.(2021)Liu, Lin, Cao, Hu, Wei, Zhang, Lin, and Guo]{swin}
Ze Liu, Yutong Lin, Yue Cao, Han Hu, Yixuan Wei, Zheng Zhang, Stephen Lin, and Baining Guo.
\newblock Swin transformer: Hierarchical vision transformer using shifted windows.
\newblock In \emph{ICCV}, pages 10012--10022, 2021.

\bibitem[Lu et~al.(2019)Lu, Batra, Parikh, and Lee]{pretrain4}
Jiasen Lu, Dhruv Batra, Devi Parikh, and Stefan Lee.
\newblock Vilbert: Pretraining task-agnostic visiolinguistic representations for vision-and-language tasks.
\newblock In \emph{NeurIPS}, 2019.

\bibitem[Luo et~al.(2023)Luo, Xiao, Liu, Li, Wang, Tang, Li, and Yang]{soc}
Zhuoyan Luo, Yicheng Xiao, Yong Liu, Shuyan Li, Yitong Wang, Yansong Tang, Xiu Li, and Yujiu Yang.
\newblock Soc: Semantic-assisted object cluster for referring video object segmentation.
\newblock \emph{arXiv preprint arXiv:2305.17011}, 2023.

\bibitem[Mottaghi et~al.(2014)Mottaghi, Chen, Liu, Cho, Lee, Fidler, Urtasun, and Yuille]{pascal}
Roozbeh Mottaghi, Xianjie Chen, Xiaobai Liu, Nam{-}Gyu Cho, Seong{-}Whan Lee, Sanja Fidler, Raquel Urtasun, and Alan~L. Yuille.
\newblock The role of context for object detection and semantic segmentation in the wild.
\newblock In \emph{CVPR}, 2014.

\bibitem[Qu et~al.(2022)Qu, Wu, Liu, Gong, Liang, Russakovsky, Zhao, and Wei]{qu2022siri}
Mengxue Qu, Yu Wu, Wu Liu, Qiqi Gong, Xiaodan Liang, Olga Russakovsky, Yao Zhao, and Yunchao Wei.
\newblock Siri: A simple selective retraining mechanism for transformer-based visual grounding.
\newblock In \emph{ECCV}, 2022.

\bibitem[Qu et~al.(2023)Qu, Wu, Wei, Liu, Liang, and Zhao]{qu2023learning}
Mengxue Qu, Yu Wu, Yunchao Wei, Wu Liu, Xiaodan Liang, and Yao Zhao.
\newblock Learning to segment every referring object point by point.
\newblock In \emph{Proceedings of the IEEE/CVF Conference on Computer Vision and Pattern Recognition}, 2023.

\bibitem[Radford et~al.(2021)Radford, Kim, Hallacy, Ramesh, Goh, Agarwal, Sastry, Askell, Mishkin, Clark, Krueger, and Sutskever]{clip}
Alec Radford, Jong~Wook Kim, Chris Hallacy, Aditya Ramesh, Gabriel Goh, Sandhini Agarwal, Girish Sastry, Amanda Askell, Pamela Mishkin, Jack Clark, Gretchen Krueger, and Ilya Sutskever.
\newblock Learning transferable visual models from natural language supervision.
\newblock In \emph{ICML}, 2021.

\bibitem[Ren et~al.(2023)Ren, Xu, Li, Hong, Zhang, and Chen]{chen6}
Jiyuan Ren, Yanggang Xu, Zuxin Li, Chaopeng Hong, Xiao-Ping Zhang, and Xinlei Chen.
\newblock Scheduling uav swarm with attention-based graph reinforcement learning for ground-to-air heterogeneous data communication.
\newblock In \emph{Adjunct Proceedings of the 2023 ACM International Joint Conference on Pervasive and Ubiquitous Computing \& the 2023 ACM International Symposium on Wearable Computing}, pages 670--675, 2023.

\bibitem[Ronneberger et~al.(2015)Ronneberger, Fischer, and Brox]{unet}
Olaf Ronneberger, Philipp Fischer, and Thomas Brox.
\newblock U-net: Convolutional networks for biomedical image segmentation.
\newblock In \emph{MICCAI}, 2015.

\bibitem[Tan and Le(2019)]{efficientnet}
Mingxing Tan and Quoc Le.
\newblock Efficientnet: Rethinking model scaling for convolutional neural networks.
\newblock In \emph{ICML}, pages 6105--6114, 2019.

\bibitem[Wang et~al.(2022)Wang, Chen, Cheng, Wu, Dang, and Chen]{chen1}
Haoyang Wang, Xuecheng Chen, Yuhan Cheng, Chenye Wu, Fan Dang, and Xinlei Chen.
\newblock H-swarmloc: Efficient scheduling for localization of heterogeneous mav swarm with deep reinforcement learning.
\newblock In \emph{Proceedings of the 20th ACM Conference on Embedded Networked Sensor Systems}, pages 1148--1154, 2022.

\bibitem[Wang et~al.(2017)Wang, Lu, Wang, Feng, Wang, Yin, and Ruan]{frequency1}
Lijun Wang, Huchuan Lu, Yifan Wang, Mengyang Feng, Dong Wang, Baocai Yin, and Xiang Ruan.
\newblock Learning to detect salient objects with image-level supervision.
\newblock In \emph{CVPR}, pages 136--145, 2017.

\bibitem[Wu et~al.(2019)Wu, Kirillov, Massa, Lo, and Girshick]{wu2019detectron2}
Yuxin Wu, Alexander Kirillov, Francisco Massa, Wan-Yen Lo, and Ross Girshick.
\newblock Detectron2.
\newblock \url{https://github.com/facebookresearch/detectron2}, 2019.

\bibitem[Xian et~al.(2019)Xian, Choudhury, He, Schiele, and Akata]{spnet}
Yongqin Xian, Subhabrata Choudhury, Yang He, Bernt Schiele, and Zeynep Akata.
\newblock Semantic projection network for zero- and few-label semantic segmentation.
\newblock In \emph{CVPR}, 2019.

\bibitem[Xie et~al.(2021)Xie, Wang, Yu, Anandkumar, Alvarez, and Luo]{segformer}
Enze Xie, Wenhai Wang, Zhiding Yu, Anima Anandkumar, Jose~M. Alvarez, and Ping Luo.
\newblock Segformer: Simple and efficient design for semantic segmentation with transformers.
\newblock In \emph{NIPS}, 2021.

\bibitem[Xu et~al.(2022)Xu, Mello, Liu, Byeon, Breuel, Kautz, and Wang]{groupvit}
Jiarui Xu, Shalini~De Mello, Sifei Liu, Wonmin Byeon, Thomas~M. Breuel, Jan Kautz, and Xiaolong Wang.
\newblock Groupvit: Semantic segmentation emerges from text supervision.
\newblock In \emph{CVPR}, 2022.

\bibitem[Xu et~al.(2023{\natexlab{a}})Xu, Liu, Vahdat, Byeon, Wang, and De~Mello]{odise}
Jiarui Xu, Sifei Liu, Arash Vahdat, Wonmin Byeon, Xiaolong Wang, and Shalini De~Mello.
\newblock Open-vocabulary panoptic segmentation with text-to-image diffusion models.
\newblock In \emph{CVPR}, pages 2955--2966, 2023{\natexlab{a}}.

\bibitem[Xu et~al.(2021)Xu, Zhang, Wei, Lin, Cao, Hu, and Bai]{Simbaseline}
Mengde Xu, Zheng Zhang, Fangyun Wei, Yutong Lin, Yue Cao, Han Hu, and Xiang Bai.
\newblock A simple baseline for zero-shot semantic segmentation with pre-trained vision-language model.
\newblock \emph{arXiv preprint arXiv:2112.14757}, 2021.

\bibitem[Xu et~al.(2023{\natexlab{b}})Xu, Zhang, Wei, Hu, and Bai]{san}
Mengde Xu, Zheng Zhang, Fangyun Wei, Han Hu, and Xiang Bai.
\newblock Side adapter network for open-vocabulary semantic segmentation.
\newblock In \emph{CVPR}, pages 2945--2954, 2023{\natexlab{b}}.

\bibitem[Xu et~al.(2019)Xu, Zhang, and Xiao]{frequency2}
Zhi-Qin~John Xu, Yaoyu Zhang, and Yanyang Xiao.
\newblock Training behavior of deep neural network in frequency domain.
\newblock In \emph{ICONIP}, pages 264--274, 2019.

\bibitem[Yin et~al.(2019)Yin, Gontijo~Lopes, Shlens, Cubuk, and Gilmer]{frequency3}
Dong Yin, Raphael Gontijo~Lopes, Jon Shlens, Ekin~Dogus Cubuk, and Justin Gilmer.
\newblock A fourier perspective on model robustness in computer vision.
\newblock \emph{NeurIPS}, 2019.

\bibitem[Zhang and Ding(2021)]{pmosr}
Hui Zhang and Henghui Ding.
\newblock Prototypical matching and open set rejection for zero-shot semantic segmentation.
\newblock In \emph{ICCV}, 2021.

\bibitem[Zhou et~al.(2017)Zhou, Zhao, Puig, Fidler, Barriuso, and Torralba]{ade20k}
Bolei Zhou, Hang Zhao, Xavier Puig, Sanja Fidler, Adela Barriuso, and Antonio Torralba.
\newblock Scene parsing through {ADE20K} dataset.
\newblock In \emph{CVPR}, 2017.

\end{thebibliography}
}


\end{document}